\documentclass[sigconf]{acmart}

\setcopyright{none}
\renewcommand\footnotetextcopyrightpermission[1]{}

\AtBeginDocument{%
  }

\setcopyright{acmlicensed}
\copyrightyear{2025}
\acmYear{2025}
\acmDOI{XXXXXXX.XXXXXXX}

\acmConference[EARL @ RecSys '25]{Workshop on Evaluating and Applying Recommender Systems with Large Language Models in conjunction with RecSys '25}{September, 2025}{Prague, Czech Republic}

\usepackage{bm}
\usepackage{amsfonts}
\usepackage{fancybox}

\begin{document}

\title{Short-Form Video Recommendations with Multimodal Embeddings: Addressing Cold-Start and Bias Challenges}

\author{Andrii Dzhoha$^*$}
\email{andrew.dzhoha@zalando.de}
\affiliation{%
  \institution{Zalando SE}
  \city{Berlin}
  \country{Germany}
}

\author{Katya Mirylenka$^*$}
\email{katya.mirylenka@zalando.ch}
\affiliation{%
  \institution{Zalando Switzerland AG}
  \city{Zürich}
  \country{Switzerland}
}

\author{Egor Malykh$^*$}
\email{egor.malykh@zalando.de}
\affiliation{%
  \institution{Zalando SE}
  \city{Berlin}
  \country{Germany}
}

\author{Marco-Andrea Buchmann}
\email{marco.andrea.buchmann@zalando.ch}
\affiliation{%
  \institution{Zalando Switzerland AG}
  \city{Zürich}
  \country{Switzerland}
}

\author{Francesca Catino}
\email{francesca.catino@zalando.ch}
\affiliation{%
  \institution{Zalando Switzerland AG}
  \city{Zürich}
  \country{Switzerland}
}

\thanks{$*$ These authors contributed equally to this work.}

\begin{abstract}
    In recent years, social media users have spent significant amounts of time on short-form video platforms. As a result, established platforms in other domains, such as e-commerce, have begun introducing short-form video content to engage users and increase their time spent on the platform. The success of these experiences is due not only to the content itself but also to a unique UI innovation: instead of offering users a list of choices to click, platforms actively recommend content for users to watch one at a time. This creates new challenges for recommender systems, especially when launching a new video experience. Beyond the limited interaction data, immersive feed experiences introduce stronger position bias due to the UI and duration bias when optimizing for watch-time, as models tend to favor shorter videos. These issues, together with the feedback loop inherent in recommender systems, make it difficult to build effective solutions. In this paper, we highlight the challenges faced when introducing a new short-form video experience and present our experience showing that, even with sufficient video interaction data, it can be more beneficial to leverage a video retrieval system using a fine-tuned multimodal vision-language model to overcome these challenges. This approach demonstrated greater effectiveness compared to conventional supervised learning methods in online experiments conducted on our e-commerce platform.
\end{abstract}



\maketitle

\section{Introduction}

Short-form video platforms have rapidly reshaped digital engagement, with users now spending substantial time consuming immersive, vertically-scrolled video feeds. This paradigm shift has prompted established domains, including e-commerce, to experiment with similar experiences in order to capture user attention and drive engagement. However, the success of these platforms is not solely attributed to the content itself, but also to a distinctive UI innovation: rather than presenting users with a list of options, the system actively curates and presents content one item at a time, creating a highly engaging, lean-back experience.

This interaction model introduces new challenges for recommender systems. Unlike traditional settings where users' preferences are inferred from explicit choices among many options, immersive feeds rely on implicit signals such as watch-time and scroll behavior. The sequential, single-item presentation amplifies position bias~\cite{10.1145/3583780.3615489}, as users are more likely to engage with content shown earlier in the feed. Moreover, optimizing for watch-time can introduce a strong duration bias, with models tending to favor shorter videos that are more easily completed~\cite{10.1145/3534678.3539092, 10.1145/3503161.3548428}. These biases are further reinforced by feedback loops~\cite{10.1145/3564284}, making it difficult to ensure fair and relevant recommendations~\cite{10.1145/3397271.3401431}, especially in the early stages of a new product where interaction data is limited.

Traditional recommender models trained from scratch, such as collaborative filtering and other supervised approaches~\cite{3366424.3386195, 2409.02856, 2959100.2959190}, are effective in mature platforms with plenty of representative data. However, they often struggle when available data is limited or exhibits strong biases, as is common in new product experiences. Counterfactual learning and bias correction methods have been proposed~\cite{10.1145/3018661.3018699}, but they require careful design and large-scale data, which are often unavailable in new experiences. Mitigating such biases is challenging~\cite{2412.08780}, as existing methods often fail to remain robust when data is sparse~\cite{10.1145/2911451.2911537}, can show high variance in their results~\cite{10.1145/3336191.3371783}, or are affected by interleaving biases~\cite{10.1145/3340531.3412031}. As a result, there is a growing need for approaches that leverage the generalization capabilities of foundation models and can be tailored for specific applications, such as launching a new short-form video experience on an e-commerce platform.

In this work, we highlight the challenges encountered when launching a new short-form video experience in e-commerce. Even with access to video interaction data, conventional methods can be influenced by duration and position biases, which may limit their effectiveness. To address these issues, we present a scalable retrieval system based on a multimodal vision-language model (CLIP~\cite{clip-2021}) that maps both user history and video content into a shared semantic space. This approach leverages the generalization capabilities of foundation models, enabling robust recommendations even in cold-start scenarios and outperforming conventional supervised learning methods in our context. To personalize recommendations, we apply few-shot learning to the CLIP model using interaction data from the main e-commerce catalog, specifically from users' Browse and Search activities, capturing nuanced user preferences. For evaluation, we utilize the expert visual language model Qwen as an LVLM-as-a-judge. Our online experiments demonstrated that this approach increased watch-time completion rates by over 39\%, while maintaining a balanced distribution of video duration, popularity, and watch-time.

Our main contributions are:
\begin{itemize}
    \item We discuss the unique position and duration biases in immersive short-form video feeds, explaining their impact on recommender system performance.
    \item We share practical lessons from launching a new immersive video product, highlighting real-world challenges and the limitations of standard approaches.
    \item We present a scalable multimodal retrieval method using vision-language models and LLM-guided evaluation and few-shot learning, which delivers improved personalization and relevance.
\end{itemize}

\section{Background}\label{sec:conventional-approaches}

A recency-based solution is often the preferred initial method for video feeds, as it requires minimal engineering effort and enables rapid prototyping and launch. By surfacing the latest content, it drives engagement and content discovery, particularly in dynamic environments. This approach also minimizes the introduction of biases and the impact of feedback loops, making it a strong, interpretable baseline for evaluating more advanced personalized or multimodal methods.

Once interaction data becomes available, personalization is typically introduced using a conventional two-tower architecture~\cite{3366424.3386195, 2959100.2959190, 2507.03789}, which has become standard for scalable candidate generation. These models learn separate user and item embeddings for efficient retrieval. More advanced user models could represent interaction history as a graph of fashion interests, leveraging Siamese graph neural networks~\cite{krivosheev2021business}. Given the limitations of early video interaction data, a pragmatic approach is to use non-trainable user embeddings from existing platform models, while training the video tower from scratch on video interactions. Relevance is often defined by videos achieving watch times above a 50\% threshold, following industry standards; however, this introduces a bias toward shorter videos~\cite{10.1145/3534678.3539092, 10.1145/3503161.3548428}. Combined with position bias and feedback loops, the resulting data can be challenging, especially in immersive feed experiences where position bias is amplified~\cite{10.1145/3583780.3615489}. Training typically uses dot-product similarity with sigmoid activation, optimized via binary cross-entropy loss, and evaluated using AUC and NDCG metrics.

Other approaches include reinforcement learning for optimizing retention and watch-time~\cite{10.1145/3543873.3584640}, real-time reranking~\cite{10.1145/3511808.3557065}, and systems designed for explicit video feedback~\cite{10.1145/3534678.3539130}. While these methods increase modeling complexity or require more data to train user representations, they do not necessarily resolve the challenges posed by bias in immersive video feeds.

\section{Multimodal retrieval approach}\label{sec:multimodal-retrieval}

\begin{figure}[t]
  \centering
  \includegraphics[width=\linewidth]{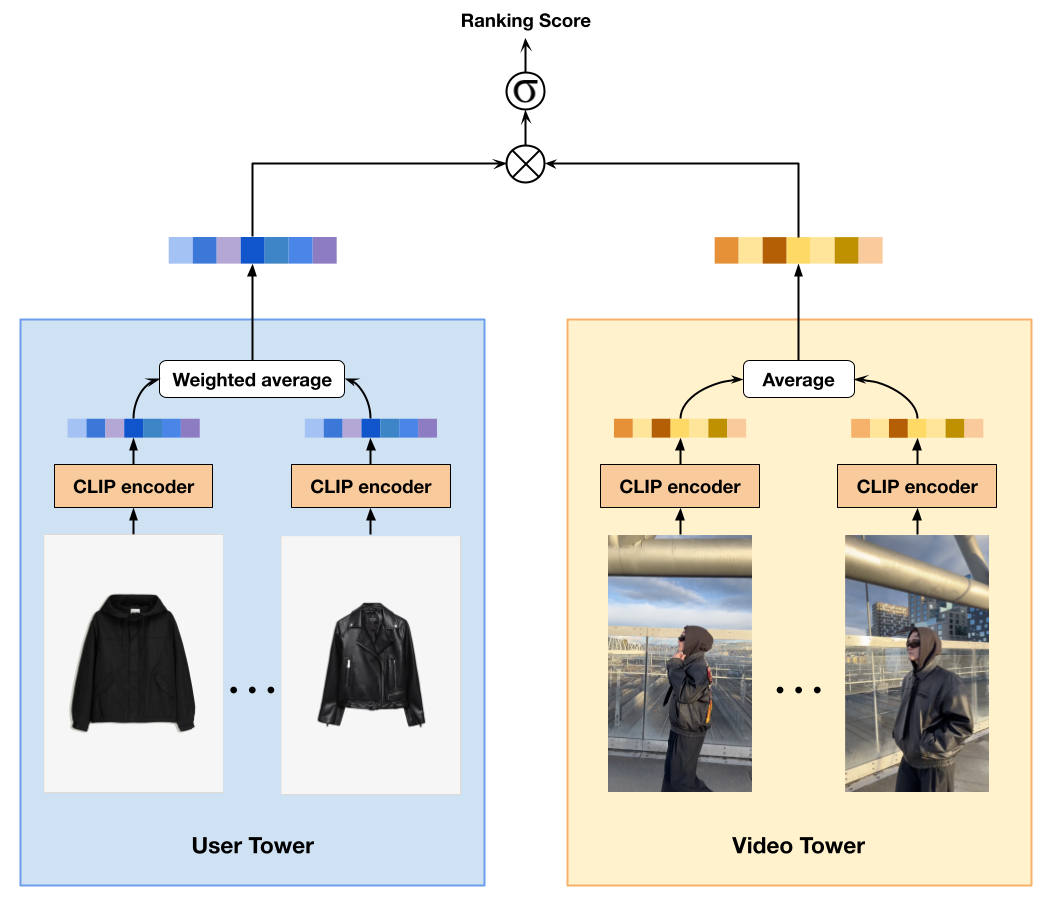}
  \caption{Overview of the multimodal retrieval architecture.}
  \label{fig:architecture}
\end{figure}

Recent advances in multimodal vision-language models, such as CLIP~\cite{clip-2021}, enable robust retrieval by embedding both users and content into a shared semantic space. We leverage this capability with a two-tower architecture to address cold-start and bias challenges in short-form video recommendations for e-commerce.

The video tower computes embeddings by averaging CLIP representations of uniformly sampled video frames, capturing visual and semantic content. The user tower aggregates CLIP embeddings of products from a user's recent interaction history, weighted by recency, to form a personalized profile. This design allows effective matching between users and videos, even without direct user-video interactions. The architecture is depicted in Figure~\ref{fig:architecture}.

A key advantage of this approach is its ability to generalize from limited data, providing meaningful recommendations in cold-start scenarios and being less susceptible to duration and position biases present in conventional models.

Further, we utilize a proprietary, adapted version of CLIP within our company, enhanced through few-shot learning on interaction data from users' Browse and Search activities. This enables us to transfer knowledge from the main catalog's interaction data via standard discriminative loss modeling, allowing the model to better approximate a relevance function for video recommendations.
Throughout this paper, all references to CLIP refer to our adapted version.

For evaluation, we employ an expert visual language model (Qwen) as an LVLM-as-a-judge, providing external relevance assessments that complement traditional metrics.

\subsection{Method}

Let $\mathcal{U}$ be the set of users and $\mathcal{V}$ the set of creator videos. For each user $u \in \mathcal{U}$, we have a time-ordered history of product interactions $H_u = [(s_1, t_1), \ldots, (s_n, t_n)]$, where $s_i$ is a product and $t_i$ is the timestamp.
Our goal is to learn a scoring function $f: \mathcal{U} \times \mathcal{V} \rightarrow \mathbb{R}$ that predicts the relevance of a video $v$ to a user $u$, producing a ranked list of videos for each user.
The scoring function is modeled as the dot product between user and video embeddings in a shared $d$-dimensional space derived from CLIP:
\[
f(u, v) = \mathbf{e}_u^\top \mathbf{e}_v,
\]
where $\mathbf{e}_u \in \mathbb{R}^{d}$ and $\mathbf{e}_v \in \mathbb{R}^{d}$ are the user and video embeddings, respectively.
The video embedding $\mathbf{e}_v$ is computed as the average of CLIP embeddings from $m$ uniformly sampled video frames:
\[
\mathbf{e}_v = \frac{1}{m} \sum_{j=1}^{m} \mathbf{E}_{\text{CLIP}}\left(\text{frame}_j\right).
\]
Video embeddings are pre-computed and indexed for efficient retrieval.
A user's embedding $\mathbf{e}_u$ is dynamically computed online as a weighted average of the CLIP embeddings of products in their recent interaction history. The CLIP product embeddings, $\mathbf{E}_{\text{CLIP}}\left(\text{s}_k\right)$, are comprehensive, incorporating all textual metadata and images associated with the product. To give more importance to recent interactions, we apply an exponential decay weighting based on the time of interaction:
\[
    \mathbf{e}_u = \frac{\sum_{k=1}^{|H_u|} w_k \mathbf{E}_{\text{CLIP}}(s_k)}{\sum_{k=1}^{|H_u|} w_k},
\]
where $w_k = \exp\left( -\lambda (t_{\text{now}} - t_k) \right)$ for decay factor $\lambda$. For new users, a global embedding based on popular products is used.

This architecture enables scalable, personalized video retrieval, with embeddings precomputed offline and user profiles computed online for real-time recommendations.

\section{Experiments}

In this section, we detail our experience personalizing the immersive short-form video feed on a large-scale e-commerce platform. The feed, designed for inspiration and entertainment, allows users to scroll through videos one at a time, similar to popular social media platforms.

\subsection{Experimental setup}

We began with a recency-based video feed to establish a baseline and collect initial user interaction data. This approach enabled rapid prototyping and provided insights into user engagement with the new experience. During this phase, we intentionally limited traffic to the new feed, allowing us to iteratively collect observations and learnings while minimizing potential risks to the broader user base.

\paragraph{VCG Conventional}%
For personalization, we first implemented a two-tower architecture. User embeddings were reused from the main e-commerce catalog model, trained on Browse and Search interactions, while the video tower was trained from scratch on video interactions. The final video embedding incorporated metadata, video ID, creator, brand, mean-pooled product and hashtag embeddings, and other relevant features, all projected through multiple non-linear layers. We framed the task as binary classification, predicting whether a video is relevant to a user, with relevance defined as watch time exceeding a 50\% threshold. Training used a contrastive loss, with positives and negatives determined by this threshold.

The main e-commerce catalog model, from which we reused user embeddings, is trained on a large sample of catalog sessions. Each session includes articles shown in response to browse or search requests, contextual data (market, device, category), user history (clicks, add-to-cart, wishlist, purchases), and subsequent interactions. The dataset comprises 250 million sessions from 70 million users across 25 markets
\footnote{Our data collection process complies with the regulations defined in the GDPR and other existing regulatory frameworks around data privacy and safety in the European Union.}.

A schematic overview of the scalable two-tower-based Video Candidate Generation (VCG) architecture is provided in Section~\ref{sec:conventional-approaches}.

\paragraph{VCG Multimodal (CLIP-based)}%
Subsequently, we introduced a multimodal retrieval system based on CLIP embeddings, as described in Section~\ref{sec:multimodal-retrieval}. This approach reused the same scalable two-tower architecture, with video embeddings precomputed and stored in an online index for efficient retrieval. User embeddings were dynamically computed from recent product interactions using CLIP representations, enabling real-time personalized video recommendations in a shared semantic space.

\subsection{Evaluation protocol}

Our evaluation procedure focuses on user engagement and is based on video feed observations. These observations are generated by the existing recency-based solution, meaning we can expect a representative sample of user preferences with fewer sampling and popularity biases that typically arise with Machine Learning models.
We use a time-based split where the test (hold-out) data is formed by sequences from the last days. This split corresponds to the actual production setup.
The ground truth is derived from the video interaction data, where relevance is modeled as a binary classification problem. A video is considered "watched" (positive example) if its watch time exceeds a 50\% threshold. Each data point represents a video feed impression, enriched with a representation of user history.

We evaluate performance primarily using feed-wise ranking metrics to compare our approach with the current recency-based method on feeds containing multiple videos. Additionally, we use video-wise binary metrics for fine-tuning solutions and measuring performance across all feeds, including single-video impressions.
\begin{enumerate}
    \item Feed-wise (list-wise) ranking metrics: NDCG, applied to a subset of the test set where each example contains at least one positive and one negative video impressions.
    \item Video-wise (point-wise) binary classification metrics: Accuracy, AUC, precision, and recall.
\end{enumerate}

To account for the position of relevant items, we weight NDCG by inverse propensity scores~\cite{10.1145/3366423.3380255}. Additionally, we monitor the skewness of popularity and watch-time distributions to assess the extent to which the model favors shorter or more popular videos~\cite{10.14778/2311906.2311916}.

The ranking metric is specifically used to assess improvements over the current recency-based production solution. The underlying assumption is that an improvement in ranking metrics over the recency-based model should correlate with an enhancement in the retrieval task, while accepting the potential bias towards more active users.

\subsection{Visual coherence evaluation}

In addition to engagement metrics, our goal is to enhance the visual appeal of the feed in relation to a user's history. To measure this, we define the visual coherence between a user and a video as the dot product of the averaged content-based embeddings of the user's past interactions and the averaged content-based embeddings of the video's associated products. This metric reflects how well a user's history aligns with a video based on content-related features such as brand, color, silhouette, and more.
These content-based embeddings are pre-extracted from product image representations, capturing visual attributes such as color, style, and silhouette to enable effective similarity comparisons.

\subsection{LVLM-as-a-judge evaluation}

To complement standard metrics, we used LVLM-as-a-judge (Large Vision-Language Model) for offline evaluation. Qwen 2.5-VL served as an external judge, rating the relevance of top-$k$ ($k=5, 10$) recommended videos based on a user's 12 most recent items of interest. Ratings were assigned on a 5-point scale, from 5 (extremely relevant) to 1 (no relevance). This approach provides an external perspective, especially valuable when user behavior data is limited. The prompt used for LVLM-as-a-judge is summarized in Figure~\ref{fig:llm-prompt}.

\begin{figure}[ht]
    \begin{center}
    \setlength{\fboxsep}{8pt}
    \colorbox{gray!10}{%
      \parbox{0.95\linewidth}{%
          {\ttfamily
            You are an AI fashion relevance analyst. Your primary function is to critically and objectively evaluate the relevance of video content against a specific user's fashion history. It is crucial that you use the defined textual relevance categories appropriately and avoid defaulting to a generally positive assessment unless there is substantial, specific evidence.\\
            <...>\\
            Assign one of the following textual categories for relevance. Choose the category that most accurately describes the alignment. Be discerning.\\
            * "excellent\_match": <...>\\
            * "good\_match": <...>\\
            * "partial\_match": <...>\\
            * "poor\_match": <...>\\
            * "no\_match": <...>
          }
      }
    }
    \end{center}
  \caption{LLM prompt for LVLM-as-a-judge evaluation.}
  \label{fig:llm-prompt}
\end{figure}

\subsection{Offline evaluation}

Offline evaluation using user behavior data and NDCG with inverse propensity scores showed only modest improvements in watch-time metrics for both VCG approaches compared to the recency baseline, with differences not reaching statistical significance. We hypothesize that position and duration biases were stronger than anticipated in the offline setting. Notably, the VCG Conventional model, trained on video interaction data, exhibited greater skewness in popularity and watch-time distributions than both the recency-based and VCG Multimodal solutions, indicating these biases affected it more. In terms of binary classification metrics, VCG Conventional achieved moderate discriminative power with AUC scores of 0.7.

\begin{figure}[t]
  \centering
  \begin{minipage}{0.48\linewidth}
    \centering
    \includegraphics[width=\linewidth]{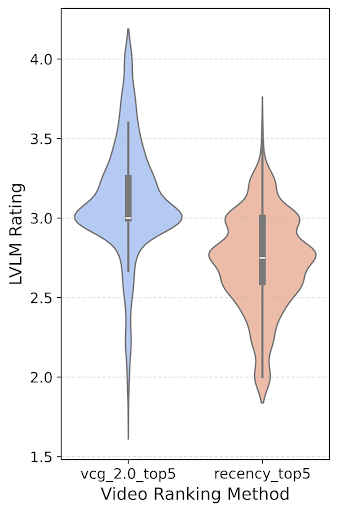}
  \end{minipage}
  \hfill
  \begin{minipage}{0.48\linewidth}
    \centering
    \includegraphics[width=\linewidth]{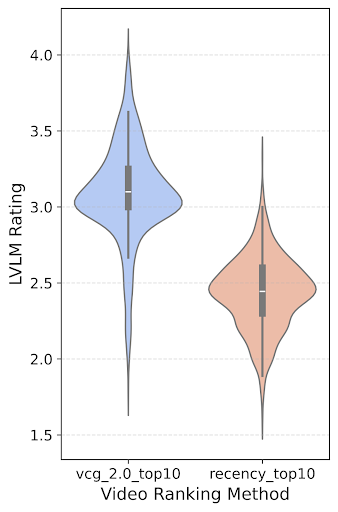}
  \end{minipage}
  \caption{Distribution of LVLM-as-a-judge scores for VCG Multimodal (CLIP-based) and recency-based baseline.}
  \label{fig:recency}
\end{figure}

Given the challenges of debiasing~\cite{2412.08780, 10.1145/3534678.3539092, 10.1145/3503161.3548428}, we shifted focus in the second iteration with VCG Multimodal to visual coherence and LVLM-as-a-judge metrics, which are especially relevant in new product scenarios with limited user interaction data. Visual coherence gains were substantial, with up to a 37\% increase for the top-100 recommended videos. LVLM-as-a-judge consistently assigned higher relevance scores to top-5 and top-10 VCG-ranked videos (see Figure~\ref{fig:recency}). These results are summarized in Table~\ref{tab:evaluation-results}, which reports absolute scores and standard deviations.

\begin{table}[t]
  \caption{
    Summary of offline evaluation: VCG Multimodal (CLIP-based) vs recency-based baseline, measured by visual coherence and average LVLM-as-a-judge scores, with standard deviation in parentheses.
  }
  \label{tab:evaluation-results}
  \centering
  \begin{tabular}{l l c c}
    \toprule
    & & \multicolumn{2}{c}{\textbf{\textbf{LVLM-as-a-judge}}} \\
    \cmidrule(lr){3-4}
    \textbf{Method} & \textbf{Visual coherence} & top-5 & top-10 \\
    \midrule
    Recency-based  & 13.8 (6.48) & 2.72 (0.30) & 2.43 (0.24) \\
    VCG Multimodal & 18.9 (8.23) & 3.12 (0.35) & 3.09 (0.32) \\
    \midrule
    Improvement &  +37\% & +14.7\% & +27\% \\
    \bottomrule
  \end{tabular}
\end{table}

\subsection{Online experiments}

We first tested the VCG Conventional approach in an online experiment, which resulted in only modest gains in engagement metrics. However, this model produced a highly skewed popularity distribution (as defined by item co-occurrence) and a strong bias toward short videos, so it was not deployed to production.

In the next iteration, we evaluated the VCG Multimodal (CLIP-based) model. This approach delivered substantial improvements: the number of videos watched for at least 25\% of their duration increased by 41\% (CI: 21\%--61\%), and those watched for at least 50\% increased by 50\% (CI: 22\%--78\%). The rates of video starts reaching 25\% and 50\% progress rose by 30\% (CI: 17\%--44\%) and 39\% (CI: 17\%--61\%), respectively. All uplifts were statistically significant, and the model maintained stable core metrics, confirming proper randomization and no negative impact on other KPIs. Importantly, popularity and duration skewness were significantly reduced compared to the VCG Conventional model. Based on these strong results, the VCG Multimodal model was rolled out to production.

\section{Conclusion}

Immersive short-form video feeds introduce unique challenges for recommender systems, particularly due to strong position and duration biases that can distort relevance and fairness -- especially in new product launches with limited interaction data. Our work demonstrates that conventional supervised approaches, while effective in mature settings, are susceptible to these biases and may not generalize well.
By leveraging a scalable multimodal retrieval system based on vision-language models and guided by LLMs for evaluation, we effectively addressed these challenges.
These results highlight the importance of foundation models and external evaluation frameworks, such as LVLM-as-a-judge, for building fair and effective recommender systems in immersive video environments.
Future work will further explore advanced LVLM-based representations to deepen user-video understanding, quantify the reliability of LLM-based judgments~\cite{bhattacharjya2025simba}, and continue mitigating bias in evolving recommendation scenarios.

\section*{Acknowledgement}

We are grateful for the valuable feedback, insightful discussions, and constant support from our many colleagues,
including Svitlana Borzenko, Ellie Langhans, Dario d'Andrea, Michael Gygli, Mahdyar Ravanbakhsh, Vahan Nanumyan, Jacek Wasilewski, Maarten Versteegh, Helen Maxwell, Dmitry Isaev, Luca Perkovic, Onur Karacali, and Juan Casero.

\balance

\bibliographystyle{ACM-Reference-Format}
\bibliography{main}


\begin{thebibliography}{22}


\ifx \showCODEN    \undefined \def \showCODEN     #1{\unskip}     \fi
\ifx \showISBNx    \undefined \def \showISBNx     #1{\unskip}     \fi
\ifx \showISBNxiii \undefined \def \showISBNxiii  #1{\unskip}     \fi
\ifx \showISSN     \undefined \def \showISSN      #1{\unskip}     \fi
\ifx \showLCCN     \undefined \def \showLCCN      #1{\unskip}     \fi
\ifx \shownote     \undefined \def \shownote      #1{#1}          \fi
\ifx \showarticletitle \undefined \def \showarticletitle #1{#1}   \fi
\ifx \showURL      \undefined \def \showURL       {\relax}        \fi
\providecommand\bibfield[2]{#2}
\providecommand\bibinfo[2]{#2}
\providecommand\natexlab[1]{#1}
\providecommand\showeprint[2][]{arXiv:#2}

\bibitem[Bhattacharjya et~al\mbox{.}(2025)]%
        {bhattacharjya2025simba}
\bibfield{author}{\bibinfo{person}{Debarun Bhattacharjya},
  \bibinfo{person}{Balaji Ganesan}, \bibinfo{person}{Junkyu Lee},
  \bibinfo{person}{Radu Marinescu}, \bibinfo{person}{Katsiaryna Mirylenka},
  \bibinfo{person}{Michael Glass}, {and} \bibinfo{person}{Xiao Shou}.}
  \bibinfo{year}{2025}\natexlab{}.
\newblock \showarticletitle{{SIMBA UQ}: Similarity-Based Aggregation for
  Uncertainty Quantification in Large Language Models}. In
  \bibinfo{booktitle}{\emph{Findings of the Association for Computational
  Linguistics: EMNLP 2025}}.
\newblock


\bibitem[Cai et~al\mbox{.}(2023)]%
        {10.1145/3543873.3584640}
\bibfield{author}{\bibinfo{person}{Qingpeng Cai}, \bibinfo{person}{Shuchang
  Liu}, \bibinfo{person}{Xueliang Wang}, \bibinfo{person}{Tianyou Zuo},
  \bibinfo{person}{Wentao Xie}, \bibinfo{person}{Bin Yang},
  \bibinfo{person}{Dong Zheng}, \bibinfo{person}{Peng Jiang}, {and}
  \bibinfo{person}{Kun Gai}.} \bibinfo{year}{2023}\natexlab{}.
\newblock \showarticletitle{Reinforcing User Retention in a Billion Scale Short
  Video Recommender System}. In \bibinfo{booktitle}{\emph{Companion Proceedings
  of the ACM Web Conference 2023}} (Austin, TX, USA)
  \emph{(\bibinfo{series}{WWW '23 Companion})}. \bibinfo{publisher}{Association
  for Computing Machinery}, \bibinfo{address}{New York, NY, USA},
  \bibinfo{pages}{421–426}.
\newblock
\showISBNx{9781450394192}
\href{https://doi.org/10.1145/3543873.3584640}{doi:\nolinkurl{10.1145/3543873.3584640}}


\bibitem[Celikik et~al\mbox{.}(2024)]%
        {2409.02856}
\bibfield{author}{\bibinfo{person}{Marjan Celikik}, \bibinfo{person}{Jacek
  Wasilewski}, \bibinfo{person}{Ana~Peleteiro Ramallo}, \bibinfo{person}{Alexey
  Kurennoy}, \bibinfo{person}{Evgeny Labzin}, \bibinfo{person}{Danilo Ascione},
  \bibinfo{person}{Tural Gurbanov}, \bibinfo{person}{Géraud~Le Falher},
  \bibinfo{person}{Andrii Dzhoha}, {and} \bibinfo{person}{Ian Harris}.}
  \bibinfo{year}{2024}\natexlab{}.
\newblock \bibinfo{title}{Building a Scalable, Effective, and Steerable Search
  and Ranking Platform}.
\newblock
\showeprint[arxiv]{2409.02856}~[cs.IR]
\urldef\tempurl%
\url{https://arxiv.org/abs/2409.02856}
\showURL{%
\tempurl}


\bibitem[Chen et~al\mbox{.}(2023)]%
        {10.1145/3564284}
\bibfield{author}{\bibinfo{person}{Jiawei Chen}, \bibinfo{person}{Hande Dong},
  \bibinfo{person}{Xiang Wang}, \bibinfo{person}{Fuli Feng},
  \bibinfo{person}{Meng Wang}, {and} \bibinfo{person}{Xiangnan He}.}
  \bibinfo{year}{2023}\natexlab{}.
\newblock \showarticletitle{Bias and Debias in Recommender System: A Survey and
  Future Directions}.
\newblock \bibinfo{journal}{\emph{ACM Trans. Inf. Syst.}} \bibinfo{volume}{41},
  \bibinfo{number}{3}, Article \bibinfo{articleno}{67} (\bibinfo{date}{feb}
  \bibinfo{year}{2023}), \bibinfo{numpages}{39}~pages.
\newblock
\showISSN{1046-8188}
\href{https://doi.org/10.1145/3564284}{doi:\nolinkurl{10.1145/3564284}}


\bibitem[Covington et~al\mbox{.}(2016)]%
        {2959100.2959190}
\bibfield{author}{\bibinfo{person}{Paul Covington}, \bibinfo{person}{Jay
  Adams}, {and} \bibinfo{person}{Emre Sargin}.}
  \bibinfo{year}{2016}\natexlab{}.
\newblock \showarticletitle{Deep Neural Networks for YouTube Recommendations}.
  In \bibinfo{booktitle}{\emph{Proceedings of the 10th ACM Conference on
  Recommender Systems}} (Boston, Massachusetts, USA)
  \emph{(\bibinfo{series}{RecSys '16})}. \bibinfo{publisher}{Association for
  Computing Machinery}, \bibinfo{address}{New York, NY, USA},
  \bibinfo{pages}{191–198}.
\newblock
\showISBNx{9781450340359}
\href{https://doi.org/10.1145/2959100.2959190}{doi:\nolinkurl{10.1145/2959100.2959190}}


\bibitem[Dzhoha et~al\mbox{.}(2024)]%
        {2412.08780}
\bibfield{author}{\bibinfo{person}{Andrii Dzhoha}, \bibinfo{person}{Alexey
  Kurennoy}, \bibinfo{person}{Vladimir Vlasov}, {and} \bibinfo{person}{Marjan
  Celikik}.} \bibinfo{year}{2024}\natexlab{}.
\newblock \bibinfo{title}{Reducing Popularity Influence by Addressing Position
  Bias}.
\newblock
\showeprint[arxiv]{2412.08780}~[cs.IR]
\urldef\tempurl%
\url{https://arxiv.org/abs/2412.08780}
\showURL{%
\tempurl}


\bibitem[Dzhoha et~al\mbox{.}(2025)]%
        {2507.03789}
\bibfield{author}{\bibinfo{person}{Andrii Dzhoha}, \bibinfo{person}{Alisa
  Mironenko}, \bibinfo{person}{Vladimir Vlasov}, \bibinfo{person}{Maarten
  Versteegh}, {and} \bibinfo{person}{Marjan Celikik}.}
  \bibinfo{year}{2025}\natexlab{}.
\newblock \bibinfo{title}{Efficient and Effective Query Context-Aware
  Learning-to-Rank Model for Sequential Recommendation}.
\newblock
\showeprint[arxiv]{2507.03789}~[cs.IR]
\urldef\tempurl%
\url{https://arxiv.org/abs/2507.03789}
\showURL{%
\tempurl}


\bibitem[Ge et~al\mbox{.}(2020)]%
        {10.1145/3397271.3401431}
\bibfield{author}{\bibinfo{person}{Yingqiang Ge}, \bibinfo{person}{Shuya Zhao},
  \bibinfo{person}{Honglu Zhou}, \bibinfo{person}{Changhua Pei},
  \bibinfo{person}{Fei Sun}, \bibinfo{person}{Wenwu Ou}, {and}
  \bibinfo{person}{Yongfeng Zhang}.} \bibinfo{year}{2020}\natexlab{}.
\newblock \showarticletitle{Understanding Echo Chambers in E-commerce
  Recommender Systems}. \bibinfo{pages}{2261--2270}.
\newblock
\href{https://doi.org/10.1145/3397271.3401431}{doi:\nolinkurl{10.1145/3397271.3401431}}


\bibitem[Gong et~al\mbox{.}(2022)]%
        {10.1145/3511808.3557065}
\bibfield{author}{\bibinfo{person}{Xudong Gong}, \bibinfo{person}{Qinlin Feng},
  \bibinfo{person}{Yuan Zhang}, \bibinfo{person}{Jiangling Qin},
  \bibinfo{person}{Weijie Ding}, \bibinfo{person}{Biao Li},
  \bibinfo{person}{Peng Jiang}, {and} \bibinfo{person}{Kun Gai}.}
  \bibinfo{year}{2022}\natexlab{}.
\newblock \showarticletitle{Real-time Short Video Recommendation on Mobile
  Devices}. In \bibinfo{booktitle}{\emph{Proceedings of the 31st ACM
  International Conference on Information \& Knowledge Management}} (Atlanta,
  GA, USA) \emph{(\bibinfo{series}{CIKM '22})}. \bibinfo{publisher}{Association
  for Computing Machinery}, \bibinfo{address}{New York, NY, USA},
  \bibinfo{pages}{3103–3112}.
\newblock
\showISBNx{9781450392365}
\href{https://doi.org/10.1145/3511808.3557065}{doi:\nolinkurl{10.1145/3511808.3557065}}


\bibitem[Joachims et~al\mbox{.}(2017)]%
        {10.1145/3018661.3018699}
\bibfield{author}{\bibinfo{person}{Thorsten Joachims}, \bibinfo{person}{Adith
  Swaminathan}, {and} \bibinfo{person}{Tobias Schnabel}.}
  \bibinfo{year}{2017}\natexlab{}.
\newblock \showarticletitle{Unbiased Learning-to-Rank with Biased Feedback}. In
  \bibinfo{booktitle}{\emph{Proceedings of the Tenth ACM International
  Conference on Web Search and Data Mining}} (Cambridge, United Kingdom)
  \emph{(\bibinfo{series}{WSDM '17})}. \bibinfo{publisher}{Association for
  Computing Machinery}, \bibinfo{address}{New York, NY, USA},
  \bibinfo{pages}{781–789}.
\newblock
\showISBNx{9781450346757}
\href{https://doi.org/10.1145/3018661.3018699}{doi:\nolinkurl{10.1145/3018661.3018699}}


\bibitem[Krivosheev et~al\mbox{.}(2021)]%
        {krivosheev2021business}
\bibfield{author}{\bibinfo{person}{Evgeny Krivosheev}, \bibinfo{person}{Mattia
  Atzeni}, \bibinfo{person}{Katsiaryna Mirylenka}, \bibinfo{person}{Paolo
  Scotton}, \bibinfo{person}{Christoph Miksovic}, {and} \bibinfo{person}{Anton
  Zorin}.} \bibinfo{year}{2021}\natexlab{}.
\newblock \showarticletitle{Business entity matching with siamese graph
  convolutional networks}. In \bibinfo{booktitle}{\emph{Proceedings of the AAAI
  Conference on Artificial Intelligence}}.
\newblock


\bibitem[Lin et~al\mbox{.}(2022)]%
        {10.1145/3534678.3539130}
\bibfield{author}{\bibinfo{person}{Zihan Lin}, \bibinfo{person}{Hui Wang},
  \bibinfo{person}{Jingshu Mao}, \bibinfo{person}{Wayne~Xin Zhao},
  \bibinfo{person}{Cheng Wang}, \bibinfo{person}{Peng Jiang}, {and}
  \bibinfo{person}{Ji-Rong Wen}.} \bibinfo{year}{2022}\natexlab{}.
\newblock \showarticletitle{Feature-aware Diversified Re-ranking with
  Disentangled Representations for Relevant Recommendation}. In
  \bibinfo{booktitle}{\emph{Proceedings of the 28th ACM SIGKDD Conference on
  Knowledge Discovery and Data Mining}} (Washington DC, USA)
  \emph{(\bibinfo{series}{KDD '22})}. \bibinfo{publisher}{Association for
  Computing Machinery}, \bibinfo{address}{New York, NY, USA},
  \bibinfo{pages}{3327–3335}.
\newblock
\showISBNx{9781450393850}
\href{https://doi.org/10.1145/3534678.3539130}{doi:\nolinkurl{10.1145/3534678.3539130}}


\bibitem[Liu et~al\mbox{.}(2023)]%
        {10.1145/3583780.3615489}
\bibfield{author}{\bibinfo{person}{Qingyun Liu}, \bibinfo{person}{Zhe Zhao},
  \bibinfo{person}{Liang Liu}, \bibinfo{person}{Zhen Zhang},
  \bibinfo{person}{Junjie Shan}, \bibinfo{person}{Yuening Li},
  \bibinfo{person}{Shuchao Bi}, \bibinfo{person}{Lichan Hong}, {and}
  \bibinfo{person}{Ed~H. Chi}.} \bibinfo{year}{2023}\natexlab{}.
\newblock \showarticletitle{Multitask Ranking System for Immersive Feed and No
  More Clicks: A Case Study of Short-Form Video Recommendation}. In
  \bibinfo{booktitle}{\emph{Proceedings of the 32nd ACM International
  Conference on Information and Knowledge Management}} (Birmingham, United
  Kingdom) \emph{(\bibinfo{series}{CIKM '23})}. \bibinfo{publisher}{Association
  for Computing Machinery}, \bibinfo{address}{New York, NY, USA},
  \bibinfo{pages}{4709–4716}.
\newblock
\showISBNx{9798400701245}
\href{https://doi.org/10.1145/3583780.3615489}{doi:\nolinkurl{10.1145/3583780.3615489}}


\bibitem[Ovaisi et~al\mbox{.}(2020)]%
        {10.1145/3366423.3380255}
\bibfield{author}{\bibinfo{person}{Zohreh Ovaisi}, \bibinfo{person}{Ragib
  Ahsan}, \bibinfo{person}{Yifan Zhang}, \bibinfo{person}{Kathryn Vasilaky},
  {and} \bibinfo{person}{Elena Zheleva}.} \bibinfo{year}{2020}\natexlab{}.
\newblock \showarticletitle{Correcting for Selection Bias in Learning-to-rank
  Systems}. In \bibinfo{booktitle}{\emph{Proceedings of The Web Conference
  2020}} (Taipei, Taiwan) \emph{(\bibinfo{series}{WWW '20})}.
  \bibinfo{publisher}{Association for Computing Machinery},
  \bibinfo{address}{New York, NY, USA}, \bibinfo{pages}{1863–1873}.
\newblock
\showISBNx{9781450370233}
\href{https://doi.org/10.1145/3366423.3380255}{doi:\nolinkurl{10.1145/3366423.3380255}}


\bibitem[Radford et~al\mbox{.}(2021)]%
        {clip-2021}
\bibfield{author}{\bibinfo{person}{Alec Radford}, \bibinfo{person}{Jong~Wook
  Kim}, \bibinfo{person}{Chris Hallacy}, \bibinfo{person}{Aditya Ramesh},
  \bibinfo{person}{Gabriel Goh}, \bibinfo{person}{Sandhini Agarwal},
  \bibinfo{person}{Girish Sastry}, \bibinfo{person}{Amanda Askell},
  \bibinfo{person}{Pamela Mishkin}, \bibinfo{person}{Jack Clark},
  {et~al\mbox{.}}} \bibinfo{year}{2021}\natexlab{}.
\newblock \showarticletitle{Learning transferable visual models from natural
  language supervision}. In \bibinfo{booktitle}{\emph{International conference
  on machine learning}}. PmLR, \bibinfo{pages}{8748--8763}.
\newblock


\bibitem[Saito et~al\mbox{.}(2020)]%
        {10.1145/3336191.3371783}
\bibfield{author}{\bibinfo{person}{Yuta Saito}, \bibinfo{person}{Suguru
  Yaginuma}, \bibinfo{person}{Yuta Nishino}, \bibinfo{person}{Hayato Sakata},
  {and} \bibinfo{person}{Kazuhide Nakata}.} \bibinfo{year}{2020}\natexlab{}.
\newblock \showarticletitle{Unbiased Recommender Learning from
  Missing-Not-At-Random Implicit Feedback}. In
  \bibinfo{booktitle}{\emph{Proceedings of the 13th International Conference on
  Web Search and Data Mining}} (Houston, TX, USA) \emph{(\bibinfo{series}{WSDM
  '20})}. \bibinfo{publisher}{Association for Computing Machinery},
  \bibinfo{address}{New York, NY, USA}, \bibinfo{pages}{501–509}.
\newblock
\showISBNx{9781450368223}
\href{https://doi.org/10.1145/3336191.3371783}{doi:\nolinkurl{10.1145/3336191.3371783}}


\bibitem[Vardasbi et~al\mbox{.}(2020)]%
        {10.1145/3340531.3412031}
\bibfield{author}{\bibinfo{person}{Ali Vardasbi}, \bibinfo{person}{Harrie
  Oosterhuis}, {and} \bibinfo{person}{Maarten de Rijke}.}
  \bibinfo{year}{2020}\natexlab{}.
\newblock \showarticletitle{When Inverse Propensity Scoring does not Work:
  Affine Corrections for Unbiased Learning to Rank}. In
  \bibinfo{booktitle}{\emph{Proceedings of the 29th ACM International
  Conference on Information \& Knowledge Management}} (Virtual Event, Ireland)
  \emph{(\bibinfo{series}{CIKM '20})}. \bibinfo{publisher}{Association for
  Computing Machinery}, \bibinfo{address}{New York, NY, USA},
  \bibinfo{pages}{1475–1484}.
\newblock
\showISBNx{9781450368599}
\href{https://doi.org/10.1145/3340531.3412031}{doi:\nolinkurl{10.1145/3340531.3412031}}


\bibitem[Wang et~al\mbox{.}(2016)]%
        {10.1145/2911451.2911537}
\bibfield{author}{\bibinfo{person}{Xuanhui Wang}, \bibinfo{person}{Michael
  Bendersky}, \bibinfo{person}{Donald Metzler}, {and} \bibinfo{person}{Marc
  Najork}.} \bibinfo{year}{2016}\natexlab{}.
\newblock \showarticletitle{Learning to Rank with Selection Bias in Personal
  Search}. In \bibinfo{booktitle}{\emph{Proceedings of the 39th International
  ACM SIGIR Conference on Research and Development in Information Retrieval}}
  (Pisa, Italy) \emph{(\bibinfo{series}{SIGIR '16})}.
  \bibinfo{publisher}{Association for Computing Machinery},
  \bibinfo{address}{New York, NY, USA}, \bibinfo{pages}{115–124}.
\newblock
\showISBNx{9781450340694}
\href{https://doi.org/10.1145/2911451.2911537}{doi:\nolinkurl{10.1145/2911451.2911537}}


\bibitem[Yang et~al\mbox{.}(2020)]%
        {3366424.3386195}
\bibfield{author}{\bibinfo{person}{Ji Yang}, \bibinfo{person}{Xinyang Yi},
  \bibinfo{person}{Derek Zhiyuan~Cheng}, \bibinfo{person}{Lichan Hong},
  \bibinfo{person}{Yang Li}, \bibinfo{person}{Simon Xiaoming~Wang},
  \bibinfo{person}{Taibai Xu}, {and} \bibinfo{person}{Ed~H. Chi}.}
  \bibinfo{year}{2020}\natexlab{}.
\newblock \showarticletitle{Mixed Negative Sampling for Learning Two-tower
  Neural Networks in Recommendations}. In \bibinfo{booktitle}{\emph{Companion
  Proceedings of the Web Conference 2020}} (Taipei, Taiwan)
  \emph{(\bibinfo{series}{WWW '20})}. \bibinfo{publisher}{Association for
  Computing Machinery}, \bibinfo{address}{New York, NY, USA},
  \bibinfo{pages}{441–447}.
\newblock
\showISBNx{9781450370240}
\href{https://doi.org/10.1145/3366424.3386195}{doi:\nolinkurl{10.1145/3366424.3386195}}


\bibitem[Yin et~al\mbox{.}(2012)]%
        {10.14778/2311906.2311916}
\bibfield{author}{\bibinfo{person}{Hongzhi Yin}, \bibinfo{person}{Bin Cui},
  \bibinfo{person}{Jing Li}, \bibinfo{person}{Junjie Yao}, {and}
  \bibinfo{person}{Chen Chen}.} \bibinfo{year}{2012}\natexlab{}.
\newblock \showarticletitle{Challenging the long tail recommendation}.
\newblock \bibinfo{journal}{\emph{Proc. VLDB Endow.}} \bibinfo{volume}{5},
  \bibinfo{number}{9} (\bibinfo{date}{may} \bibinfo{year}{2012}),
  \bibinfo{pages}{896–907}.
\newblock
\showISSN{2150-8097}
\href{https://doi.org/10.14778/2311906.2311916}{doi:\nolinkurl{10.14778/2311906.2311916}}


\bibitem[Zhan et~al\mbox{.}(2022)]%
        {10.1145/3534678.3539092}
\bibfield{author}{\bibinfo{person}{Ruohan Zhan}, \bibinfo{person}{Changhua
  Pei}, \bibinfo{person}{Qiang Su}, \bibinfo{person}{Jianfeng Wen},
  \bibinfo{person}{Xueliang Wang}, \bibinfo{person}{Guanyu Mu},
  \bibinfo{person}{Dong Zheng}, \bibinfo{person}{Peng Jiang}, {and}
  \bibinfo{person}{Kun Gai}.} \bibinfo{year}{2022}\natexlab{}.
\newblock \showarticletitle{Deconfounding Duration Bias in Watch-time
  Prediction for Video Recommendation}. In
  \bibinfo{booktitle}{\emph{Proceedings of the 28th ACM SIGKDD Conference on
  Knowledge Discovery and Data Mining}} (Washington DC, USA)
  \emph{(\bibinfo{series}{KDD '22})}. \bibinfo{publisher}{Association for
  Computing Machinery}, \bibinfo{address}{New York, NY, USA},
  \bibinfo{pages}{4472–4481}.
\newblock
\showISBNx{9781450393850}
\href{https://doi.org/10.1145/3534678.3539092}{doi:\nolinkurl{10.1145/3534678.3539092}}


\bibitem[Zheng et~al\mbox{.}(2022)]%
        {10.1145/3503161.3548428}
\bibfield{author}{\bibinfo{person}{Yu Zheng}, \bibinfo{person}{Chen Gao},
  \bibinfo{person}{Jingtao Ding}, \bibinfo{person}{Lingling Yi},
  \bibinfo{person}{Depeng Jin}, \bibinfo{person}{Yong Li}, {and}
  \bibinfo{person}{Meng Wang}.} \bibinfo{year}{2022}\natexlab{}.
\newblock \showarticletitle{DVR: Micro-Video Recommendation Optimizing
  Watch-Time-Gain under Duration Bias}. In
  \bibinfo{booktitle}{\emph{Proceedings of the 30th ACM International
  Conference on Multimedia}} (Lisboa, Portugal) \emph{(\bibinfo{series}{MM
  '22})}. \bibinfo{publisher}{Association for Computing Machinery},
  \bibinfo{address}{New York, NY, USA}, \bibinfo{pages}{334–345}.
\newblock
\showISBNx{9781450392037}
\href{https://doi.org/10.1145/3503161.3548428}{doi:\nolinkurl{10.1145/3503161.3548428}}


\end{thebibliography}

\end{document}